\documentclass{llncs}
\usepackage{graphicx}

\begin{document}

\title{A Computational Model of Spatial Memory Anticipation during Visual Search}

\author{J{\'e}r{\'e}my Fix and Julien Vitay and Nicolas P. Rougier}

\institute{Loria, Campus Scientifique, BP239 \\
54506 Vandoeuvre-les-Nancy, France}

\maketitle

\begin{abstract}
  Some visual search tasks require to memorize the location of stimuli
  that have been previously scanned. Considerations about the eye movements raise the
  question of how we are able to maintain a coherent memory, despite
  the frequent drastically changes in the perception. In this
  article, we present a computational model that is able to anticipate
  the consequences of the eye movements on the visual perception in
  order to update a spatial memory.
\end{abstract}

\section{Introduction}

While the notion of anticipation has been known for quite a long time
in both psychology, biology or physics domains, it remains difficult
to agree on a standard definition that can account for its multiple
facets. For example, in \cite{Grush2004}, the author proposes an
analogy between motor control and kalman filters where a controller is
supposed to produce a signal that is sent to both the plant to control
and to the emulator that is then able to produce a prediction of the
behavior. In \cite{Riegler2001}, the author refutes this standard
definition of anticipatory systems as being based on a predictive
model of the system itself and its environment.\\


However, even if there does not exist such a general definition, there
is a large consensus on the fundamental role played by anticipation in
behavior. Someone that would have been deprived from any anticipation
abilities would be severely impaired in its everyday life, from both a
perception and action point of view. Of course, the deprivation of any
anticipatory capabilities does not need to be so radical and we can
also imagine a lighter impairment of the system. For instance, let us
simply consider the inability to anticipate changes in the visual
information resulting from an eye saccade. This anticipation is known
to be largely based on unconscious mechanisms that provide
us with a feeling of stability while the whole retina is submerged by
different information at each saccade : producing a
saccade results in a complete change in the visual perception of the
outer world. If a system is unable to anticipate its own saccadic
movements, it cannot pretend to obtain a coherent view of the world:
each image would be totally uncorrelated from the others. One stimulus
being at one location before a saccade could not be identified easily
at being the same stimulus at another location after the saccade. The
aim of this paper is to precisely pinpoint the importance of this
visual anticipation in establishing a coherent view of the environment
and to propose a computational model that rely on anticipation to
efficiently scan a visual scene.\\

After a quick review of the literature demonstrating that visual
anticipation is a critical part of the visual system, we introduce a
simple experiment of visual search and explain how the model we
propose can solve the task by using both anticipation and a dynamic
model of working memory.

\section{Visual search}

Visual search is a cognitive task that most generally involves an
active scan of a visual scene for finding one or several given targets
among distractors. It is deeply anchored in most animal behaviors,
from a predator looking for a prey in the environment, to the prey
looking for a safe place to avoid being seen by the predator.
Psychological experiments may be less ecological and may propose for
example to find a given letter among an array of other letters,
measuring the efficiency of the visual search in terms of reaction
time (the average time to find the target given the experimental paradigm).
In the early eighties, \cite{Treisman1980} suggested that the brain
actually extracts some basic features from the visual field in order
to perform the search. Among these basic features that have been
recently reviewed by \cite{Wolfe1998}, one can find features such as
color, shape, motion or curvature. Finding a target is then equivalent
to finding the conjunction of features (that may be unique) that best
describ the target. In this sense, \cite{Treisman1980} distinguished
two main paradigms (a more tempered point of view can be found in
\cite{Duncan1989}).
\\

\noindent {\bf Feature search} refers to a search where the target differs from
distractors against exactly one feature.\\
{\bf Conjunction search} refers to a search where the target differs
from distractors against two or more features.
\\

\begin{figure}
\begin{center}
\includegraphics[width=8.5cm]{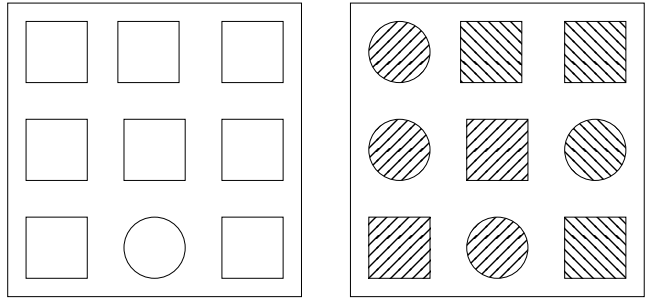}
\caption{Feature search can be performed very quickly as
  illustrated on the left part of the figure; the disc shape literally
  pops out from the scene. However, as illustrated on the right part
  of the figure, if the stimuli share at least two features, the pop out
  effect is suppressed. Hence, finding the disc shape with the stripes
  going from up-left to down-right requires an active scan of the
  visual scene. \label{fig:search}} 
\end{center}
\end{figure}

What characterizes best the feature search is a constant search time
that does not depend on the number of distractors. The target is
sufficiently different from the distractors to pop out. However, in the case of conjunction search, the time to
find the target seems to be tightly linked to the number of distractors
that share at least one feature with the target (cf. Fig.
\ref{fig:search}). These observations lead to the question of how a visual stimulus could be
represented in the brain. In \cite{Milner1992}, the authors proposed that the visual
perception relies on two separated pathways: one would be dedicated to
the extraction of features independently on their spatial positions
(the so-called {\em What} pathways) while the other would only extract
stimuli position without any information regarding feature properties
(the so-called {\em Where} pathway). In this article, we don't deal with the
high-level processing of the visual input (the {\em What} pathway) nor with the difficult problem of
the communication between the two pathways known as the binding
problem and only consider a spatial representation of the visual
input, filled by computing basic filters.

\subsection{Saccadic eye movements}

The eye movements may have different behavioral goals, leading to five different
categories of movements : saccades, vestibulo-ocular reflex,
optokinetic reflex, smooth-pursuit and vergence. However, in this article
we will only focus on saccades (for a detailed study of eye movements,
see \cite{Leigh1999}, \cite{Carpenter1988}).\\

Saccades are fast and frequent eye movements that move quickly the eye
from the current point of gaze to a new location in order to center a
visual stimulus on the fovea, a small area on the retina where the
resolution is at its highest. The velocity of the eyes
depends on the amplitude of the movement and can be reached up to 700
degrees per second at a frequency of 3 Hz. The question we would like
to address is how the brain may give the illusion of a stable visual
space while the visual perception is drastically modified every
200~ms.\\

While the debate to decide whether or not the brain is blind during a
saccade has not been settled (\cite{Kleiser2004}, \cite{Ross2001}),
the coherence between the perception before and after a saccade cannot be established
accurately solely based on perception. One solution is to consider
that the brain may use an efferent copy of the voluntary eye movement
to remap the representation it has built of the visual world. Several
studies shed light on pre-saccadic activities in areas such as V4 and
LIP where the locations of relevant stimuli are supposed to be
represented. In \cite{Moore1998}, the authors suggest that ``the
presaccadic enhancement exhibited by V4 neurons [...] provides a
mechanism by which a clear perception of the saccade goal can be
maintained during the execution of the saccade, perhaps for the
purpose of establishing continuity across eye movements''. In
\cite{Merriam2005}, the authors review evidences that LIP neurons,
whose receptive field will land on a previously stimulated screen
location after a saccade, are excited even if the stimulus disappears
during the saccade.

\subsection{Visual attention}

The capacity to focus on a given stimulus of the visual scene is
tightly linked to visual attention that has been defined as the
capacity to concentrate cognitive ressources on a restricted subset of
sensory information (\cite{James1890}). In the context of visual
attention, only a small subset of the retina information is available at any given
time to elaborate motor plans or cognitive reasoning (cf.
\emph{change blindness} experiments presented in \cite{Regan2001},
\cite{Simons2000}). The selection of a target for an eye movement is
then closely related to the notion of spatial attention
(\cite{Moore2001}) that is classically divided into two types:
{\bfseries overt attention} which involves a saccade to center an object
on the fovea and {\bfseries covert attention} in which no eye movement is
initiated.  These two types of spatial attention were first supposed
to be independent (\cite{Posner1990}) but recent studies such as the
premotor theory of attention proposed in \cite{Rizzolatti1987} (see
also \cite{Chelazzi1993}, \cite{Kowler1995}, \cite{Craighero1999})
consider that covert and overt attention rely on the same neural
structures but movement is inhibited in covert attention.

\subsection{Computational models}
\label{section:computational}

Over the past few years, several attempts at modeling visual
attention have been engaged (\cite{Ullman1985}, \cite{Tsotsos1995},
\cite{Wolfe2000}, \cite{Itti2001}, \cite{Hamker2004}). The basic idea
behind most of those models is to find a way to select interesting
locations in the visual space giving their behavioral relevance and
whether or not they have been already focused. The two central notions
in this context have been proposed by \cite{Ullman1985} and \cite{Posner1984}:
\begin{itemize}
\item saliency map
\item inhibition of return (IOR).
\end{itemize}
The saliency map is a single spatial map, in retinotopic coordinates, where all the available visual
information converge in order to obtain a unified representation of
stimuli, according to their behavioral relevances. A winner-take-all algorithm can be easily used to find what
is the most salient stimulus within the visual scene which is
identified as the attentional point of focus. However, in order
to be able to go to the next stimuli, it is important to bias the
winner-take-all algorithm in such a way that it prevents returning to
an already focused stimulus. The goal of the inhibition of return
mechanism is precisely to feed the saliency map with such a bias. The
idea is to have another neural map that records focused stimuli and
inhibits the corresponding locations in the saliency map. Since an
already focused stimulus is actively inhibited by this map, it cannot
pretend to win the winner-take-all competition, even if it is the most
salient.\\

The existence of a single saliency map is still not proved. In
\cite{Hamker2004} the author proposes a more distributed representation of these
relevances, clearly dividing the what and the where pathways stated
before, and where spatial competition occurs in a motor map instead of
a perceptive one. The related model exhibits good performances
regarding visual search task in natural scene, but is restricted to
covert attention. Therefore, authors do not take into account eye
movements and the visual scene is supposed to remain stable: scanning
is done without any saccade. During the rest of this article, we will
stick to the saliency map hypothesis, even if controverted, in order
to illustrate the anticipatory mechanism.

\section{A model of visual search with overt attention}

\subsection{Experiment}

In order to accurately evaluate the model, we setup a simple
experimental framework where some identical stimuli are drawn on a
blackboard and are observed by a camera. The task is to successively
focus (i.e. center) each one of the stimuli without focusing twice on
any of them. We estimate the performance of the model in terms
of how many times a stimulus has been focused. Hence, the point is not
to analyze the strategy of deciding which stimulus has to be focused
next (see \cite{Findlay2006a,Findlay2006b} for details on this
matter). In the context of the proposed model, the strategy is simply
to go from the most salient stimulus to the least salient one, and to
randomly pick one stimulus if the remaining ones are equally salient.

\begin{figure}
\begin{center}
\includegraphics[width=8.5cm]{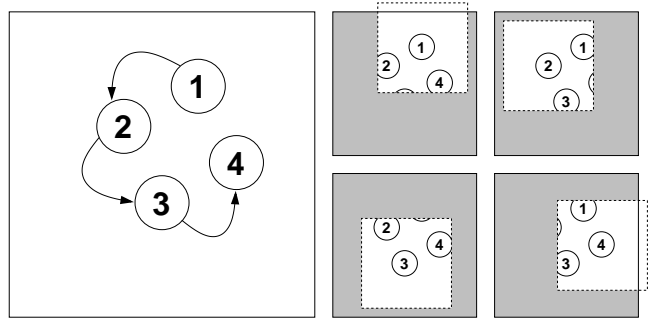}
\caption{When scanning a visual scene, going for example from
  stimulus 1 to stimulus 4, as illustrated on the left of the figure,
  the image received on the retina is radically changed when each
  stimulus is centered on the retina, as illustrated on the right of
  the figure. The difficulty in this situation is to be able to
  remember which stimulus has already been centered in order
  to center another one. The figures on the stimuli are shown only for
  explanation purpose and do not appear on the screen; all the stimuli
  are identical. \label{fig:scan}} 
\end{center}
\end{figure}

Figure \ref{fig:scan} illustrates an experiment composed of four
identical stimuli where the visual scan path has been materialized.
The effect of making a saccade from one stimulus to another is shown
and underlines the difficulty (for a computational model) of
identifying a stimulus before and after a saccade. Each one of the
stimulus being identical to the others, it is impossible to perform an
identification based solely on features.

\subsection{Model}
\begin{figure}
\begin{center}
\includegraphics[width=8.5cm]{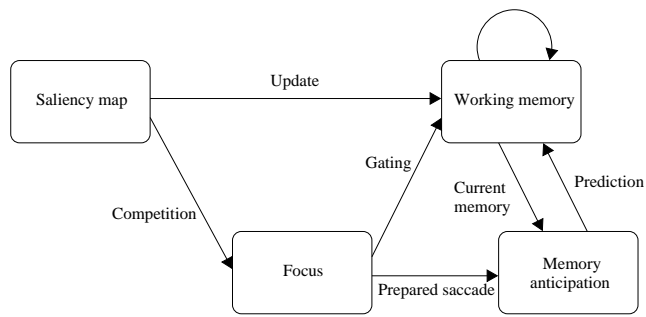}
\caption{Schematic view of the architecture of the
  model. The image captured by the camera is filtered and represented in the
  saliency map. This information feeds two pathways : one to the
  memory and one to the focus map. A competition in the focus map
  leads to the most salient location that is the target for the next
  saccade. The anticipation circuit predicts the future state of the
  memory with its current content and the programmed saccade. \label{fig:SchemaModele}} 
\end{center}
\end{figure}

The model is based on three distinct mechanisms (cf. Fig.
\ref{fig:SchemaModele} for a schematic view of the model). The first
one is a competition mechanism that involves potential targets
represented in a saliency map that were previously computed according
to visual input. Second, to be able to focus only once on each
stimulus, the locations of the scanned targets are stored in a memory map
using retinotopic coordinates. Finally, since we are considering overt
attention, the model is required to produce a camera movement,
centering the target onto the fovea, used to update the working
memory. This third mechanism works in conjunction with two inputs:
current memory and parameters of the next saccade. This allows the
model to compute quite accurately a prediction of the future state of
the visual space, restricted to the targets that have been already
memorized. A different version of this model, without the
anticipatory mechanism can be found in \cite{Vitay2005}.\\

Moreover, the model uses the computational paradigm of two dimensional
discrete neural fields (the mathematical basis of this paradigm can be
found in \cite{Amari1977} for the one dimensional case, extended to a two
dimensional study in \cite{Taylor1999}). The model consists of six
{\em n$\times$n} maps of units, characterized by their position in a
map, denoted {\bfseries x} $\in [1..n]^2$ and their activity as a function
of their position and time, denoted u({\bfseries x},t). The basic dynamic
equation that follows the activity of a unit at position {\bfseries x},
depends on its input, computed as a weighted sum over input units, and
on an weighted influence of the lateral units in the same map. Equation (\ref{equation_cnft}) is the equation proposed in
\cite{Amari1977}, discretized in space, where M is the set of the
lateral units, $M~'$ the set of the input units, $w_{M}(x-x~')$ the lateral
connection weight function, and $s(x,y)$ the afferent connection weight
function. Usually, the weighting functions $s(x,y)$ and $w_{M}(x-x~')$ are chosen
as a Gaussian or as a difference of Gaussians, as given by (\ref{function_weight}).

\begin{eqnarray}
\tau.\frac{\partial u(x,t)}{\partial t} & = & -u(x,t) + \sum_{\mathrm{M}}
w_{\mathrm{M}}(x-x~')u(x~',t) + \sum_{\mathrm{M~'}} s(x,y).u(y,t) 
\label{equation_cnft} \\
\nonumber s(x,y) & = & C.e^{\frac{{\|x-y\|^2}}{c^2}} \mbox{ with } C,c \in
   \bbbr^{*+} \\
w_{\mathrm{M}} (x-x~') & = &
A.e^{\frac{\|x-x~'\|^2}{a^2}}-B.e^{\frac{\|x-x~'\|^2}{b^2}} \mbox{ with } A,B,a,b \in \bbbr^{*+} \label{function_weight}
\end{eqnarray}

where $u(\textbf{x},t)$ is the activity of the unit at the location {\bfseries x}
in a map M, $u(\textbf{x~'},t)$ the activity of the unit at the location
{\bfseries x~'} in the same map, u(y,t) the activity of the unit at the
location {\bfseries y} in a map M~', different from M and $\tau$ is a
given parameter that defines the temporal dynamics . A unit whose
activity satisfies (\ref{equation_cnft}) will be called a
sigma unit in the following. We also introduce sigma-pi units
(\cite{Rumelhart1987}) whose
activity satisfies (\ref{equation_sigmapi}). While in
(\ref{equation_cnft}) the input of a unit is computed as a
sum of activities, in (\ref{equation_sigmapi}), the input of
the unit is computed as a sum of product of activities.

\begin{equation}
\tau.\frac{\partial u(x,t)}{\partial t} =  -u(x,t) + \sum_{\mathrm{M}}
w_{\mathrm{M}}(x-x~')u(x~',t) + \sum_{i \in \mathrm{I}} w_{i}.\prod_{y \in \mathrm{M_{i}~'}} u(y,t) 
\label{equation_sigmapi}
\end{equation}

In the following, we denote $I(${\bfseries x}$,t)$ the input of the unit {\bfseries x}, at time t,
that can be written as :

\begin{eqnarray}
I(x,t) & = & \sum_{\mathrm{M~'}} s(x,y).u(y,t) \mbox{  for sigma units} \\
I(x,t) & = & \sum_{i \in \mathrm{I}} w_{i}.\prod_{y \in \mathrm{M_{i}~'}} u(y,t) \mbox{  for
  sigma-pi units}
\end{eqnarray}

We will now describe briefly how the different maps interact. Since
the scope of this article is the anticipation mechanism, the
description of the saliency map, the focus map and the working memory
will not be accurate but a more detailed explanation, with the
appropriate dynamical equations, can be found in \cite{Vitay2005}.

\subsubsection{Saliency map}
The saliency map is updated by convolving the image captured with the
camera of the robot used for the simulation with gaussian filters. The
stimuli we use are easily discriminable from the background on the
basis of the color information. This computation leads to a
representation of the visual stimuli with gaussian patterns of
activity in a single saliency map. We point out again that this is one
of our working hypothesis, detailed in section
\ref{section:computational}.

\subsubsection{Focus}
Units in the focus map have direct excitatory feedforward inputs from
the saliency map. The lateral connections are locally excitatory and
widely inhibitory so that a competition between the units within the
map leads to the emergence of only one stimulus in the focus map. This
stimulus is the next target to focus and the movement to perform to
center it on the fovea is decoded from this map.

\subsubsection{Working memory}
Once a stimulus has appeared within the focus map and because it is
also present in the saliency map, it emerges immediately within the
working memory. Both excitations from the focus map and the saliency
map (at a same location) are necessary for the emergence of the stimulus
in the working memory area. If the focused stimulus changes, it will not
be present anymore in the focus map such that an additional mechanism
is needed to maintain it in the memory. It is not shown on the
schematic illustration \ref{fig:SchemaModele} but the memory consists
in two maps that share excitatory connections in the two ways : the
first map excites the second and the second excites the first,
weighted so that the excitation is limited in space.

\subsubsection{Memory anticipation}
The memory anticipation mechanism aims at predicting what should be
the state of the working memory, after an eye movement needed to
center the stimulus in the focus map, before the movement is
initiated. The sigma-pi units in the anticipation map has two inputs : the
activity of the units of the focus map and the activity of the units
of the working memory. If we denote wm({\bfseries x},t) the activity of
the unit {\bfseries x} of the working memory at time t, and
f({\bfseries x},t) the activity of the unit {\bfseries x} of the
focus map at time t, we define the input I({\bfseries x}) of the unit {\bfseries x} in
the anticipation map as :

\begin{equation}
I(\textbf{x},t) = \beta.\sum_{\textbf{y} \in \bbbr^2}
wm(\textbf{y},t).f(\textbf{y}-\textbf{x},t)
\label{poids_anticipation}
\end{equation}

The input of each unit in the anticipation map is computed as a
convolution product of the working memory and the focus, centered on
its coordinates. To make (\ref{poids_anticipation}) clearer, the condition of the sum is weaker than the
one that should be used : since the input maps are discrete sets of units,
the two vectors {\bfseries y} and {\bfseries y}-{\bfseries x} mustn't exceed
the size of the maps.\\
From (\ref{equation_sigmapi}) and
(\ref{poids_anticipation}), the activity of the units in the
anticipation map, without lateral connections, satisfies (\ref{equation_anticipation}).

\begin{equation}
\tau.\frac{\partial u(x,t)}{\partial t} =  -u(x,t) + \beta.\sum_{\textbf{y} \in \bbbr^2}
wm(\textbf{y},t).f(\textbf{y}-\textbf{x},t)
\label{equation_anticipation}
\end{equation}

Then, the shape of activity in the anticipation map converges to the
convolution product of the working memory and the focus map. Since the
activity in the focus map has a gaussian shape and the working memory
can be written as a sum of gaussian functions, the convolution product
of the working memory and the focus map leads to an activity profile
that is the profile in the working memory translated by the vector
represented in the focus map. This profile is the prediction of the
future state of the working memory and is then used to slightly excite
the working memory. After the eye movement and when the saliency map
is updated, the previously scanned stimuli emerge in the working
memory as a result of the conjunction of the visual stimuli in the
saliency map and the prediction of the working memory; This is the
same mechanism than the one used when a stimulus emerges in the
working memory owing to the conjunction of the activity in the
saliency map and the focus map.

\subsection{Simulation and results}
\begin{figure}
  \begin{minipage}{0.85\linewidth}
    \centering
    \begin{tabular}[htbp]{cccc}
      \includegraphics[width=0.25\linewidth]{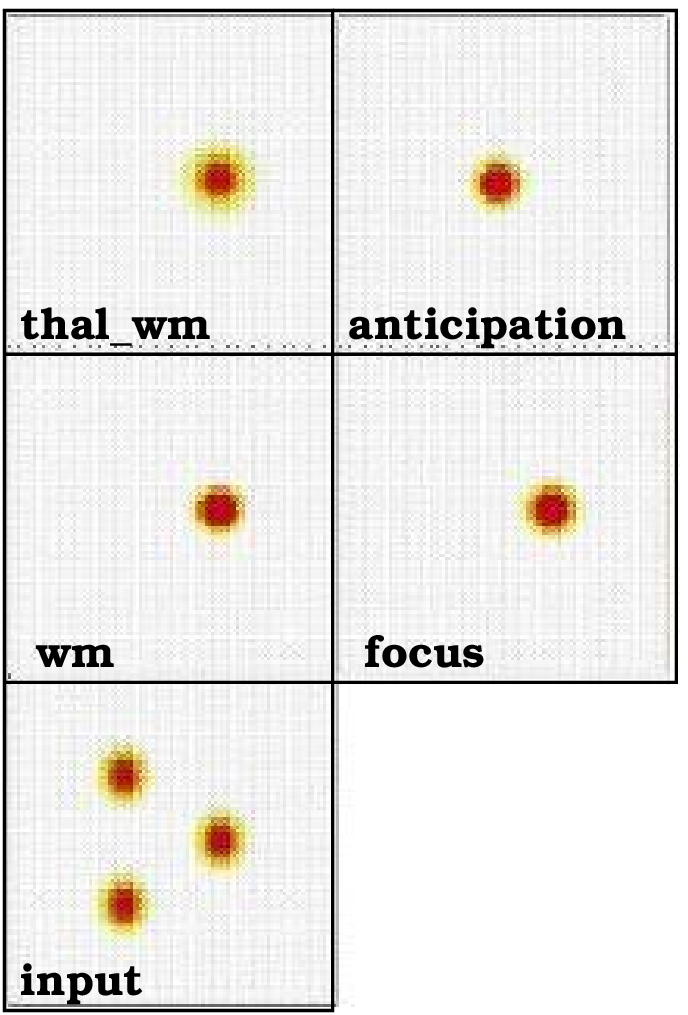}
      &
      \includegraphics[width=0.25\linewidth]{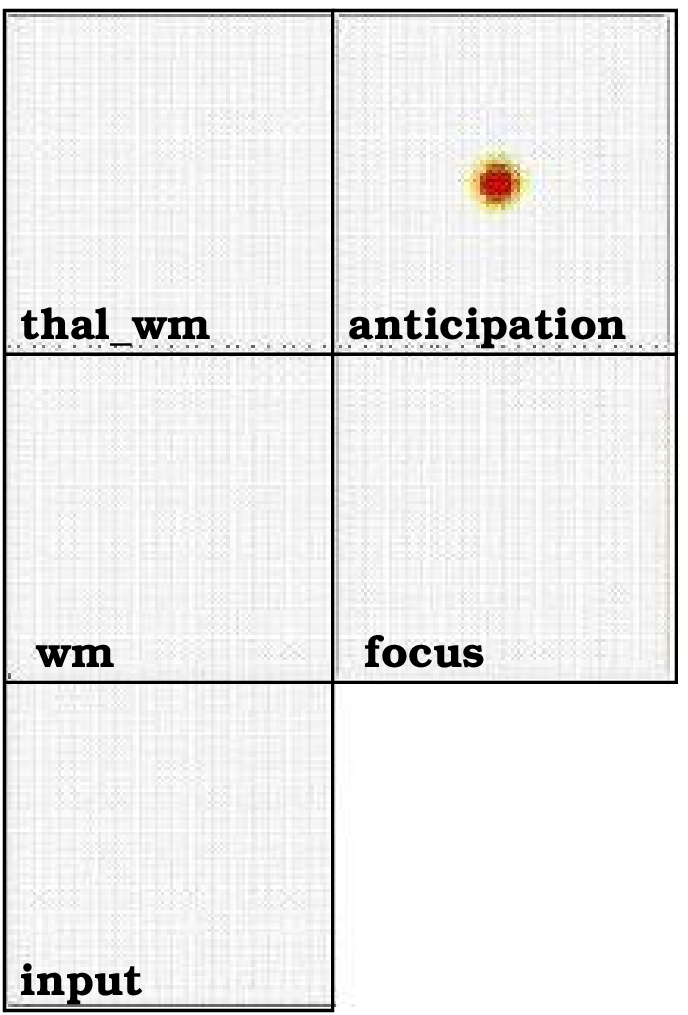}
      &
      \includegraphics[width=0.25\linewidth]{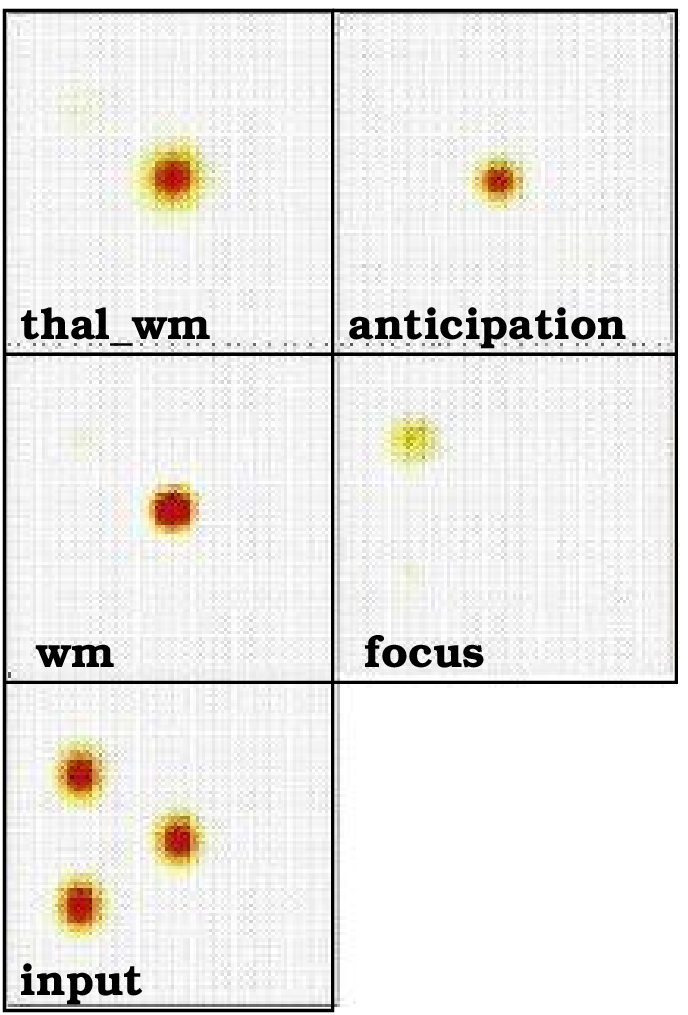}
      &
      \includegraphics[width=0.25\linewidth]{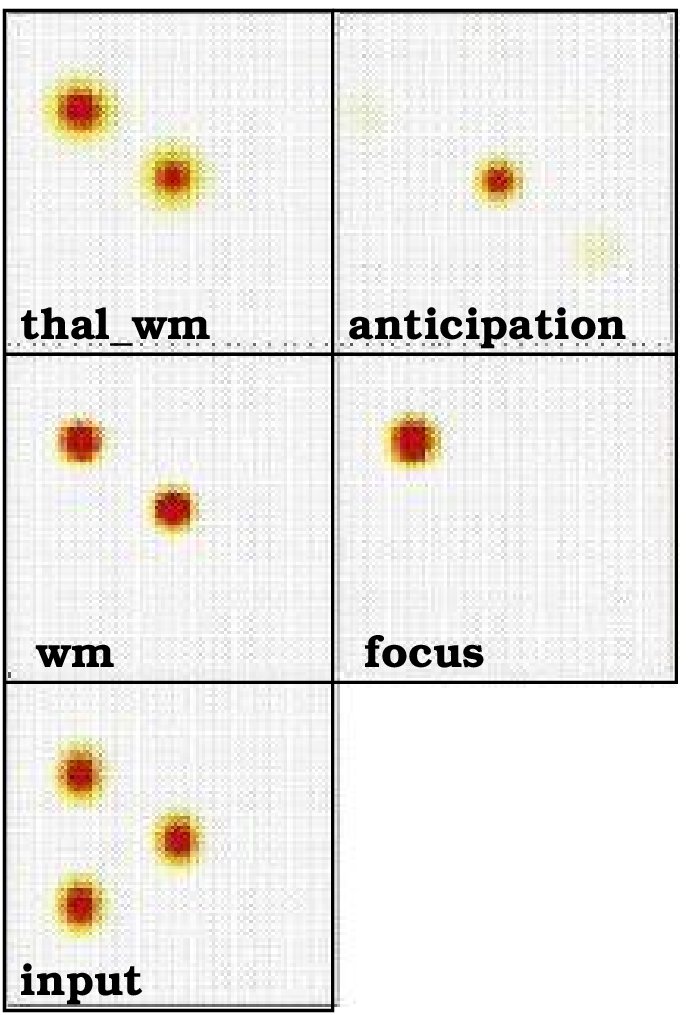}\\
      a) & b) & c) & d)\\
      \includegraphics[width=0.25\linewidth]{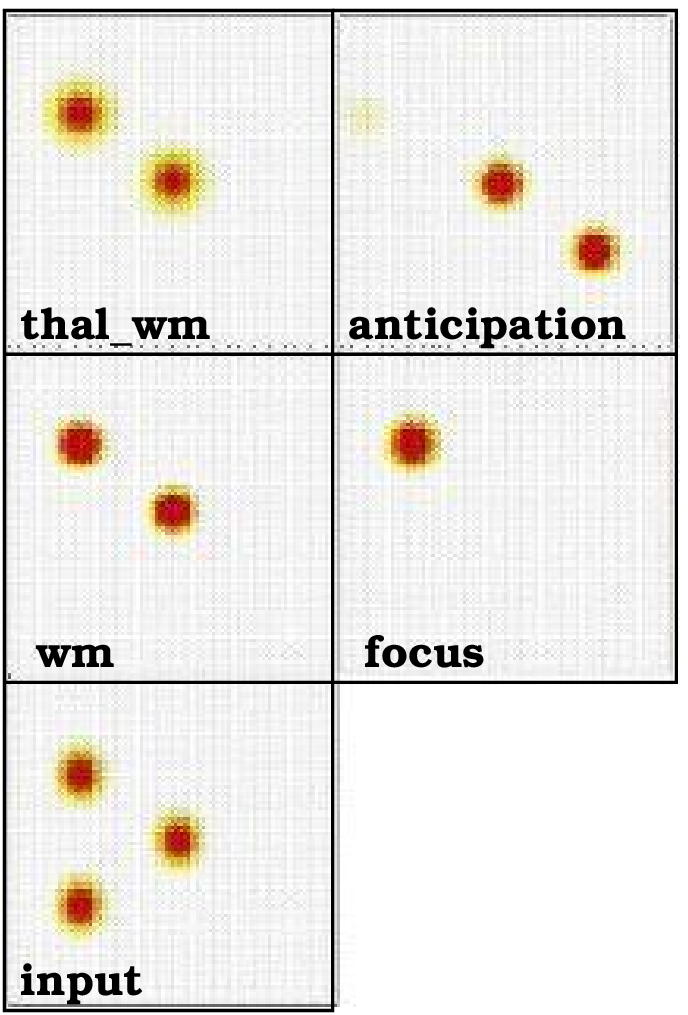}
      &
      \includegraphics[width=0.25\linewidth]{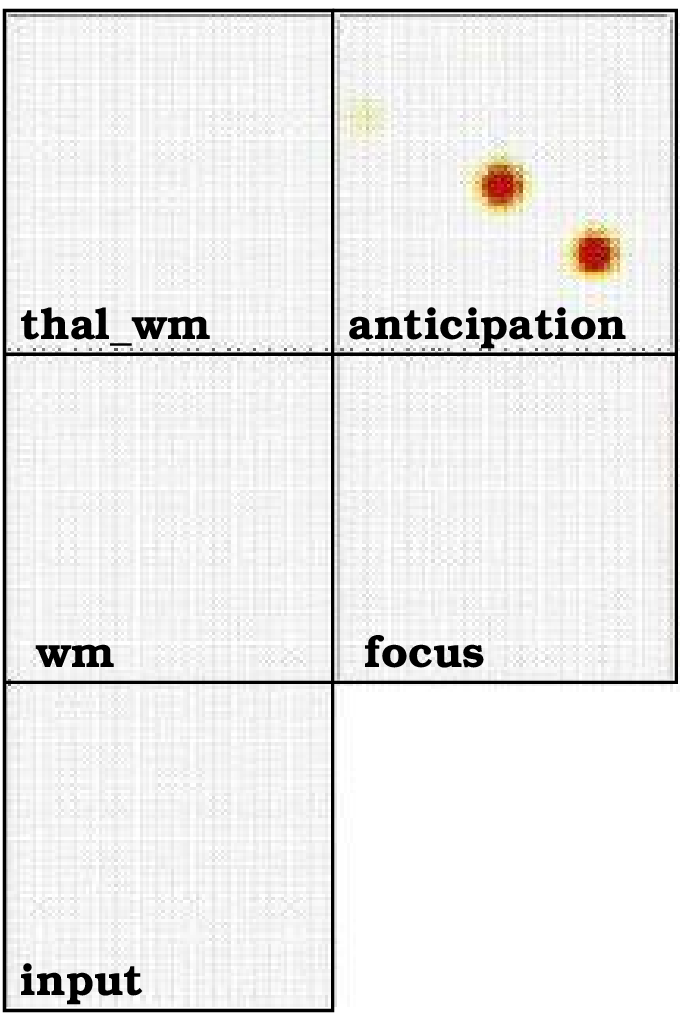}
      &
      \includegraphics[width=0.25\linewidth]{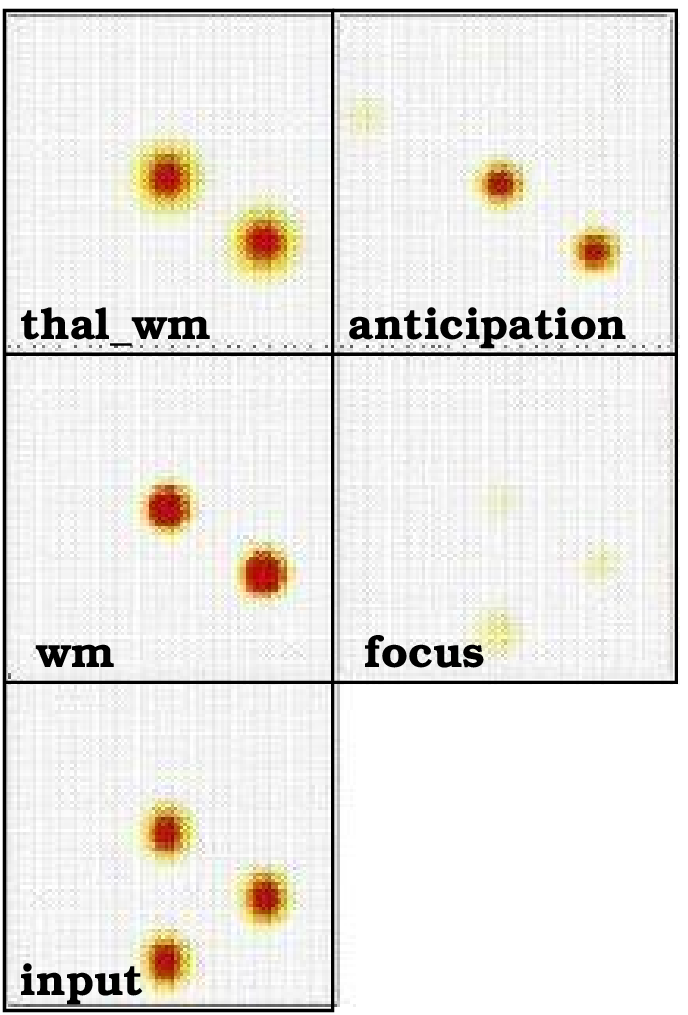}
      &
      \includegraphics[width=0.25\linewidth]{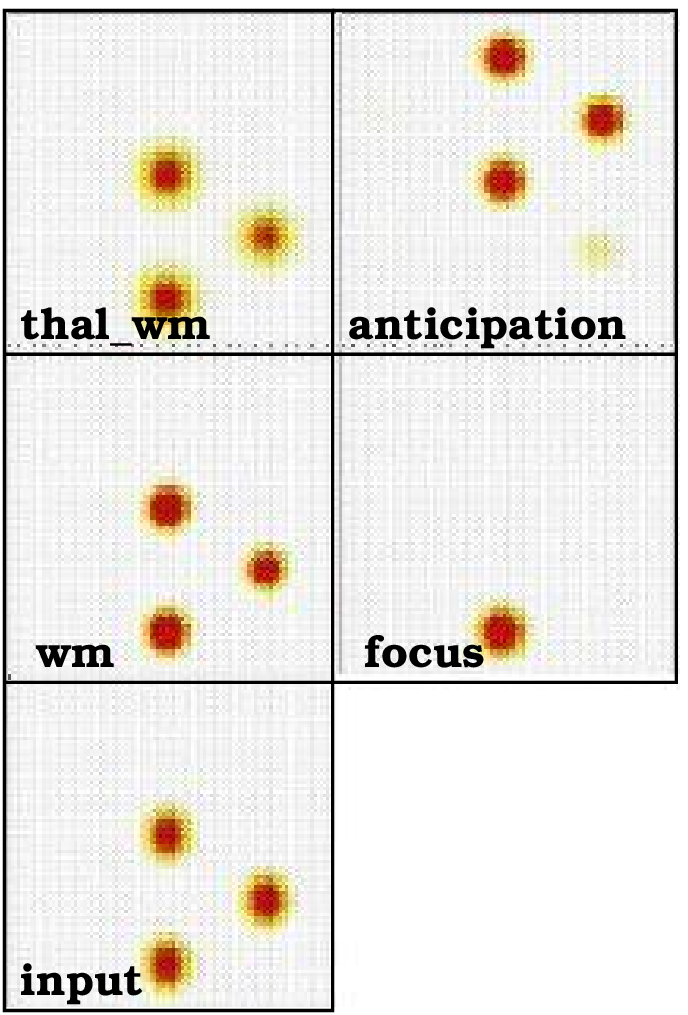}\\
      e) & f) & g) & h)\\
    \end{tabular}
  \end{minipage}
  \caption {A sequence of evolution of the model during an overt visual
    scan trial. a) One of the three stimuli emerges in the focus map and
    the anticipation's units predict the future state of the visual
    memory (the maps wm and thal\_wm). b) During the execution of the
    saccade, only the units in the anticipation map remain active. c)
    The focused stimulus emerge in the memory since it is both in the
    saliency map and the anticipation map at the same location. d) A new
    target to focus is elicited. e) The future state of the memory is
    anticipated. f) The saccade is executed and only the prediction
    remains. g) The two already focused stimuli emerge in the memory. h)
    The attentional focus lands on the last target.\label{Simulation_EPS}}
\end{figure}

The visual environment consists in three identical stimuli that the
robot is expected to scan successively exactly once. A stimulus
is easily discriminable from the background, namely a green lemon on a
white table. A complete activation sequence of the different maps is
illustrated on Fig. \ref{Simulation_EPS}. The saliency map is filled
by convolving the image captured from the camera by a green filter in
HSV coordinates such that it leads to three distinct stimuli.

At the beginning of the simulation (Fig. \ref{Simulation_EPS}a),
only one of the three stimuli emerges in the focus map, thanks to the
strong lateral competition that occurs within this map. This stimulus,
present both in the focus map and in the saliency map, emerges in the working
memory. The activation within the
anticipation map reflects what should be the state of the saliency
map, restricted to the stimuli that are in the working memory, after
the movement that brings the focused one in the center of the visual
field. During the eye movement (Fig. \ref{Simulation_EPS}b), no
visual information is available and the parameter $\tau$ in \ref{equation_cnft} and \ref{equation_anticipation} is adjusted so that only the units in the
anticipation map remain active, whereas the activity of the others
tends to zero. After the eye movement and as soon as the saliency
map is fed with the new visual input, the working memory is updated thanks to the excitation
from both saliency and anticipation map at a same location : the
prediction of the state of the visual memory is compared with the
current visual information. A new target can now be elicited in the
focus map thanks to a switch mechanism similar to that described in \cite{Vitay2005}.

\section{Discussion}

We have presented a computational model of visual memory anticipation
that is able to ensure the coherence of the visual world despite
abrupt changes in the perception that occur after each eye movement.
The prediction of the future state of the visual memory enriches the
perception of the visual world in order to avoid focusing twice a same
stimulus. As we explained previously, saccades are generally too fast
and it is impossible, even in the case we were not blind during eye
movements, to continuously update a visual memory.  An efferent copy
of the eye movement is used to establish the missing link between the
pre and post-saccadic perceptions. This mechanism is clearly an
extension of visual attention models that have been presented in
section \ref{section:computational} and where the visual world is
purely static.\\

The question of learning the underlying transformation of the
anticipatory mechanism, namely the convolution product of the focus
map and the working memory, remains open and still studied. We did
implement a learning mechanism, under restrictions and strong
hypotheses, that relies heavily on the difference between the
pre-saccadic prediction and the post-saccadic actual perception. This
self generated signal is able to measure to what extent the
predicition is correct or not. Hence, it is quite easy to modify
weights accordingly. The main difficulty during learning remains the
sampling distribution of examples within the input space which is a
well known problem in information and learning theory. Without an
additional motivational system that could bias examples according to a
given task, it is quite unrealistic to rely on a regular distribution
of examples.\\


\bibliographystyle{splncs} 
\bibliography{biblio}

\begin{thebibliography}{10}

\bibitem{Grush2004}
Grush, R.:
\newblock The emulation theory of representation : motor control, imagery and
  perception.
\newblock Behavioral and brain sciences \textbf{27} (2004)  377--442

\bibitem{Riegler2001}
Riegler, A.:
\newblock The role of anticipation in cognition.
\newblock Computing Anticipatory Systems: CASYS 2000 - Fourth International
  Conference \textbf{573} (2001)  534--541

\bibitem{Treisman1980}
Treisman, A., Gelade, G.:
\newblock A feature-integration theory of attention.
\newblock Cognitive Psychology \textbf{12}(1) (1980)  97--136

\bibitem{Wolfe1998}
Wolfe, J.:
\newblock Visual search.
\newblock In: Attention, University College London Press (1998)

\bibitem{Duncan1989}
Duncan, J., Humphreys, G.:
\newblock Visual search and stimulus similarity.
\newblock Psychological Review \textbf{96}(3) (1989)  433--458

\bibitem{Milner1992}
Goodale, M., Milner, A.:
\newblock Seperate visual pathways for perception and action.
\newblock Trends in Neurosciences \textbf{15}(1) (1992)  20--25

\bibitem{Leigh1999}
Leigh, R., Zee, D.:
\newblock The neurology of eye movements, 3rd edition.
\newblock (1999)

\bibitem{Carpenter1988}
Carpenter, R.:
\newblock Movements of the eyes, 2nd edition.
\newblock (1988)

\bibitem{Kleiser2004}
Kleiser, R., Seitz, R., Krekelberg, B.:
\newblock Neural correlates of saccadic suppression in humans.
\newblock Current Biology \textbf{14} (2004)  386--390

\bibitem{Ross2001}
Ross, J., Morrone, C., Goldberg, M., Burr, D.:
\newblock Changes in visual perception at the time of saccades.
\newblock Trends in Neurosciences \textbf{24}(2) (2001)  113--121

\bibitem{Moore1998}
Moore, T., Tolias, A., Schiller, P.:
\newblock Visual representations during saccadic eye movements.
\newblock Neurobiology \textbf{95}(15) (1998)  8981--8984

\bibitem{Merriam2005}
Merriam, E., Colby, C.:
\newblock Active vision in parietal and extrastriate cortex.
\newblock The Neuroscientist \textbf{11}(5) (2005)  484--493

\bibitem{James1890}
James, W.:
\newblock The principles of psychology.
\newblock (1890)

\bibitem{Regan2001}
O'Regan, Noe:
\newblock A sensorimotor account of vision and visual consciouness.
\newblock Behavioral and Brain Sciences \textbf{24} (2001)  939--1031

\bibitem{Simons2000}
Simons, J.:
\newblock Current approaches to change blindness.
\newblock Visual Cognition \textbf{7}(1--2--3) (2000)  1--15

\bibitem{Moore2001}
Moore, T., Fallah, M.:
\newblock Control of eye movements and spatial attention.
\newblock PNAS \textbf{98}(3) (2001)  1273--1276

\bibitem{Posner1990}
Posner, M., Petersen, S.:
\newblock The attentional system of the human brain.
\newblock Annual Review of Neurosciences \textbf{13} (1990)  25--42

\bibitem{Rizzolatti1987}
Rizzolatti, G., Riggio, L., Dascola, I., Ulmita, C.:
\newblock Reorienting attention accross the horizontal and vertical meridians.
\newblock Neuropsychologia \textbf{25} (1987)  31--40

\bibitem{Chelazzi1993}
Chelazzi, L., Miller, E., Duncan, J., Desimone, R.:
\newblock A neural basis for visual search in inferior temporal cortex.
\newblock Nature \textbf{363} (1993)  345--347

\bibitem{Kowler1995}
Kowler, E., Andersen, E., Dosher, B., Blaser, E.:
\newblock The role of attention in the programming of saccade.
\newblock Vision Research \textbf{35} (1995)  1897--1916

\bibitem{Craighero1999}
Craighero, L., Fadiga, L., Rizzolatti, G., Umilta, C.:
\newblock Action for perception : a motor-visual attentional effect.
\newblock Journal of Experimental Psychology \textbf{25} (1999)  1673--1692

\bibitem{Ullman1985}
Koch, C., Ullman, S.:
\newblock Shifts in selective visual attention : Towards the underlying neural
  circuitry.
\newblock Human Neurobiology \textbf{4}(4) (1985)  219--227

\bibitem{Tsotsos1995}
Tsotsos, J., Culhane, S., Lai, W., Davis, N.:
\newblock Modeling visual attention via selective tuning.
\newblock Artificial Intelligence \textbf{78} (1995)  507--545

\bibitem{Wolfe2000}
Wolfe, J.:
\newblock Visual attention.
\newblock In: Seeing : Handbook of Perception and Cognition, 2nd ed., De Valois
  KK (2000)  335--386

\bibitem{Itti2001}
Itti, L., Koch, C.:
\newblock Computational modeling of visual attention.
\newblock Nature Reviews Neuroscience \textbf{2}(3) (2001)  194--203

\bibitem{Hamker2004}
Hamker, F.:
\newblock A dynamic model of how feature cues guide spatial attention.
\newblock Vision Research \textbf{44} (2004)  501--521

\bibitem{Posner1984}
Posner, M., Cohen, Y.:
\newblock Components of visual orienting.
\newblock (1984)  531--556

\bibitem{Findlay2006a}
Findlay, J., Brown, V.:
\newblock Eye scanning of multi-element displays: I. scanpath planning.
\newblock Vision Research \textbf{46}(1--2) (2006a)  179--195

\bibitem{Findlay2006b}
Findlay, J., Brown, V.:
\newblock Eye scanning of multi-element displays: Ii. saccade planning.
\newblock Vision Research \textbf{46}(1--2) (2006b)  216--227

\bibitem{Vitay2005}
Vitay, J., Rougier, N.:
\newblock Using neural dynamics to switch attention.
\newblock In: International Joint Conference on Neural Networks, IJCNN (2005)

\bibitem{Amari1977}
Amari, S.:
\newblock Dynamical study of formation of cortical maps.
\newblock Biological Cybernetics \textbf{27} (1977)  77--87

\bibitem{Taylor1999}
Taylor, J.:
\newblock Neural bubble dynamics in two dimensions.
\newblock Biological Cybernetics \textbf{80} (1999)  5167--5174

\bibitem{Rumelhart1987}
Rumelhart, D., Hinton, G., McClelland, J.:
\newblock A general framework for parallel distributed processing.
\newblock In: Parallel Distributed Processing, Vol. 1, MIT Press (1987)

\end{thebibliography}

\end{document}